%% file: RR-7387.tex
\documentclass[a4paper]{article}
\usepackage[utf8]{inputenc}
\usepackage[T1]{fontenc} 

\usepackage{RR}
\usepackage{hyperref}

\graphicspath{{figs/}, {./}}

\newcommand\imf{\gamma}
\newcommand\ma{m_{\mathrm{p}}}

\RRdate{September 2010}

\RRauthor{
Romain Tavenard\thanks{Corresponding author.}
\and
Laurent Amsaleg
\and
Herv\'e J\'egou
}

\title{Balancing clusters to reduce response time variability}
\authorhead{Tavenard, Amsaleg \& J\'egou}
\titlehead{Balacing clusters to reduce response time variability}

\RRetitle{Balancing clusters to reduce response time variability in large scale image search}
\RRtitle{Réduction de la variabilité du temps de réponse pour la recherche d'image}

\RRresume{
  De nombreux algorithmes de recherche approchée de plus proches voisins en
  grande dimension partitionnent les données en clusters. Au moment de la requête,
  pour éviter une recherche exhaustive coûteuse, un index sélectionne un ou plusieurs
  clusters parmis les plus proches de la requête. Les clusters sont souvent
  obtenus par la méthode du $k$-means. Un des avantages de cette méthode
  est qu'elle tend à produire des clusters de tailles diverses. Ce déséquilibre
  entre les cardinalités des clusters a un effet négatif tant sur la variance que sur 
  l'espérance du temps de réponse. Cet article propose de modifier les centroïdes 
  obtenus par $k$-means dans le but de produire des clusters de tailles comparables.
  Les expériences effectués sur une grande collection d'images décrites montrent
  que notre algorithme réduit significativement la variance du temps de réponse,
  en diminuant légèrement les performances en termes de compromis entre efficience et 
  qualité des résultats retournés.
}

\RRabstract{
  Many algorithms for approximate nearest neighbor search in
  high-dimensional spaces partition the data into
  clusters. At query time, in order to avoid exhaustive search, 
  an index selects the few (or a single) clusters nearest to the query point. Clusters
  are often produced by the well-known $k$-means approach since it has
  several desirable properties.  On the downside, it tends to
  produce clusters having quite different cardinalities. Imbalanced
  clusters negatively impact both the variance and the expectation of
  query response times. This paper proposes to modify $k$-means centroids 
  to produce clusters with more comparable
  sizes without sacrificing the desirable properties.  
  Experiments with a large scale collection of image descriptors 
  show that our algorithm
  significantly reduces the variance of response times, at a slight cost 
  with respect to the trade-off between efficiency and search quality. 
}

\RRmotcle{recherche de plus proches voisins, grandes bases de donn\'ees, 
distance euclidienne, quantification}
\RRkeyword{nearest neighbor search, large databases, quantization}

\RRprojet{Texmex}
\RRtheme{\THCog}

\RCRennes
\URRennes

\begin{document}
\RRNo{7387}
\makeRR

\input{./intro.tex}
\input{./analysis.tex}

\input{./strategy.tex}
\input{./experiments.tex}
\input{./conclusion.tex}

\input{./biblio.bbl}
\end{document}

%% file: intro.tex
\section{Introduction}
\label{sec:intro}

Finding the nearest neighbors of high-dimensional query points still
receives a lot of research attention as this fundamental process is
central to many content-based applications. Most approaches rely 
on some different kinds of partitioning of the data collection into clusters of
descriptors. At query time, an indexing structure 
selects the few (or a single) clusters nearest to the query point. Each 
candidate cluster is scanned, actual distances to its points
are computed and the query result is built upon these distances.

There are various options for clustering points, the most popular
being the $k$-means approach. Its popularity is caused by its nice
properties: it is a simple algorithm, surprisingly effective 
and easy to implement. It nicely deals with the true distribution of data
in space by minimizing the mean square error over the clustered data
collection. On the downside, it tends to produce clusters
having quite different cardinalities. This, in turn, impacts the
performance of the retrieval algorithm: scanning heavily filled
clusters is costly as the distances to many points must be
computed. In contrast, under-filled clusters are cheap to process, 
but they are selected less often as the query descriptor is also 
less likely to be associated with these less populated clusters.
Overall, having imbalanced clusters impact both the variance and the
expectation of query response times. This is very detrimental to
contexts in which performance is paramount, such as high-throughput
settings where the true resource consumption can no more be
accurately predicted by costs models.

This phenomenon has an even more detrimental impact at large scale. In
this case, clusters must be stored on disks and the performance
severely suffer when fetching large clusters due to the large
I/Os. Furthermore, $k$-means is known to fail clustering at very large
scale, and hierarchical or approximate $k$-means must be used, which,
in turn, tend to increase the imbalance between clusters~\cite{JDS10a}.

This paper proposes an extension of the traditional $k$-means
algorithm to produce clusters of much more even size. This is
beneficial to performances since it reduces the variance and the
expectation of query response times. Balancing is obtained by slightly
distorting the boundaries of clusters. This, in turn, impacts the
quality of results since clusters do not correspond 
to the initial optimization criterion anymore.  
Section~\ref{sec:analysis} defines the problem we are
addressing and introduces the key metrics later used in the
evaluation. Section~\ref{sec:strategy} details the balancing strategy
we propose. Section~\ref{sec:experiments} evaluates the impact of
balancing on the response time of queries when using large collections
of descriptors computed over 1 million images from Flickr. 
It also shows result quality remains satisfactory with respect 
to the original $k$-means. 
Section~\ref{sec:conclusion} concludes the paper.


%% file: analysis.tex
\section{Problem statement}
\label{sec:analysis}

\subsection{Base Clustering and Searching Methods}
Without loss of generality, we partition a collection of
high-dimensional feature vectors into clusters defining Voronoi
cells. We typically use a $k$-means algorithm quantizing the data into
$k$ cells. Each cell stores a list of the vectors it clusters. This
approach is widely adopted in the context of image searches, 
where clustering is applied to local~\cite{SiZ03,NiS06} 
or global descriptors~\cite{DJSAS09,JDSP10}.  A search strategy exploiting this
partitioning is usually approximate: only one or a few cells are
explored at query time. The quality of results is typically
increased when multiple cells are probed during the search as
in~\cite{LJWCL07,JB08,JDS10a,JDSP10}. The actual distances between the query
point and the features stored in each such cell are subsequently
computed~\cite{DIIM04,ML09}. Therefore, the response time of a query is directly related
to (i) the strategy used to identify the cells to explore 
and (ii) the total number of vectors used in distance
computations. The cost for (i) is fixed and mainly corresponds to
finding the $\ma$ centroids that are the closest to the query point
($L_2$). In contrast, the cost for (ii) heavily depends on the
cardinality of each cell to process. It is of course linked to
$\ma$. Note that (i) is often negligible compared to (ii).



\subsection{Metrics: Selectivity and Recall} 

All approximate nearest-neighbor search methods try to find the best
trade-off between result quality and retrieval time. The quality of
the results can be seen as the probability to retrieve the correct
neighbors at search time, given the total amount of data that is
processed. This can be expressed in terms of selectivity and recall
defined as:
\begin{itemize}
\item \emph{Selectivity} is the total rate of
  vectors used in the distance calculations (with respect to the whole
  data collection). Obviously, the larger selectivity, the more costly is
  (ii).
\item \emph{Recall} is, for a query, the total rate
  of nearest neighbors correctly identified (with respect to the above
  selectivity). This measurement is called precision in~\cite{ML09}, but
  \emph{recall} is more accurate here. Observe that if the true
  nearest neighbor is found within any of the selected cells then it
  will be ranked first in the result list.
\end{itemize}

\subsection{Imbalance Factor}
As in~\cite{JDS10a}, we measure the imbalance between the
cardinalities of the clusters resulting from a $k$-means using an
\emph{imbalance factor $\imf$} defined as:
\begin{equation}
\imf = k \sum_{i=1}^{k} {p_i}^2
\end{equation}
where $p_i$ is the probability that a given vector is stored in the list 
associated with the $i$\textsuperscript{th} cluster. 
For a fixed dataset of size~$N$, this factor is empirically measured 
based on the number~$n_i\approx p_i \, N$ of descriptors associated with each list. 
As shown in~\cite{JDS10a}, for $\ma=1$ and for a fixed~$k$, 
the measure $\imf$ of the balancing 
is directly related to the search cost: 
a measure $\imf=3$ means that the expectation of the search time 
is three times higher than the one associated with perfectly balanced clusters. 
Optimal balancing is obtained when $n_i=n_{\mathrm{opt}}=N/k$
for all $i$. In that case, $\imf=1$ (lowest possible value) 
and the variance of query time is zero, as any cell contains exactly 
the same number of elements. This clearly appears in the analytical 
expression of the variance of the number of elements in a given list:
\begin{equation}
\mathrm{Var} = N^2 \sum_{i=1}^k p_i \left(p_i - \frac{1}{k}\right)^2. 
\end{equation}


%% file: strategy.tex
\section{Balancing Clusters}
\label{sec:strategy}

\subsection{The Balancing Process}
Balancing clusters is an iterative post-processing step performed on
the final output of a $k$-means type-of algorithm. The idea is to
artificially enlarge the distances between the data points and the
centroids of the heavily filled clusters. These penalties applied to
distances depend on the population of clusters. Hence, the
contents of cells and thus their population can be recomputed
accordingly. This balancing process eventually converges to
equally filled clusters.

The penalties are called \emph{penalization terms} and are computed as 
follows:
\begin{equation}
\left\{
\begin{array}{l}
\forall i, b^{0}_{i} = 1\\
\forall l, b^{l+1}_{i} = b^{l}_{i} \left(\frac{n^{l}_{i}}{n_{\mathrm{opt}}}\right)^{\alpha}
\end{array}
\right.
\label{equ:update}
\end{equation}
where $\alpha$ controls the convergence speed. A small value for
$\alpha$ indeed ensures that balancing will be done in a smooth way, 
while it implies to iterate more in order to get even cell population.
Note that, at each iteration $l$, the populations $n^{l}_i$ are 
updated in order to take these penalization terms into account. More 
precisely, distances from any point $\mathbf{x}$ to the 
$i^{\mathrm{th}}$ centroid are computed as
\begin{equation}
d^{l}_{\mathrm{bal}}(\mathbf{x},\mathbf{c_i})^ {2} = 
d(\mathbf{x},\mathbf{c_i})^ {2} + b^{l}_{i} .
\label{equ:disbal}
\end{equation}

\begin{figure}[t!]
\centering
\includegraphics[width=\linewidth]{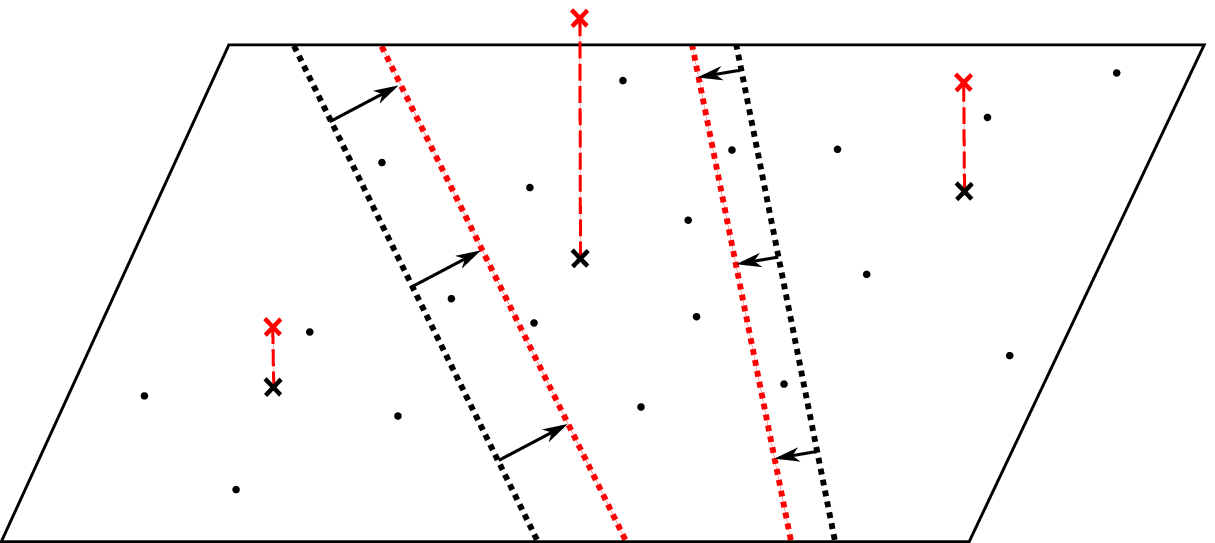}
\caption{Data points and centroids embedded in a 3-$d$ example. Data 
points are plotted as dots while centroids are represented as crosses, 
with a non-null z-axis value after some iterations.}
\label{fig:geom2d+1}
\centering
\includegraphics[width=\linewidth]{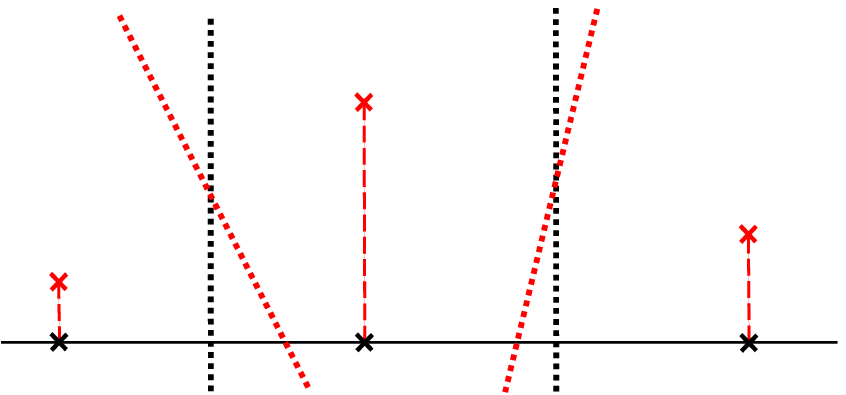}
\caption{Voronoi cell boundaries shifted after some iterations. 
New boundaries are plotted as dashed red lines which shrink the central 
cluster because of its large population.}
\label{fig:geom_borders}
\end{figure}

\subsection{Geometrical Interpretation}
\label{sec:geominterpretation}

A geometrical interpretation of the balancing process described above
is possible. Assume the balancing process first embeds the $k$-means
clustered $d$-dimensional vectors into a $(d+1)$-dimensional space. 
In this space, their $d$ first components are the ones they had in their original
space, while component $d+1$ is set to zero for all vectors. Centroids are
also embedded in the same way, except for their last component. This
last value for centroid $i$ is set to $\sqrt{b^0_i}$. Then, while the
balancing procedure iterates, it is set to the appropriate $\sqrt{b^l_i}$ value.
The intuition is that centroids are artificially elevated in an
iterative manner from the hyperplane where vectors lie. The more
vectors in one cluster, the more elevation its centroid gets.  This is
illustrated in Figure~\ref{fig:geom2d+1}, where the z-axis corresponds
to the added dimension.  Along iterations, the updated vector assignments are 
computed with respect to the coordinates of the points lying in the
augmented space. The artificial elevation of centroids tends to shrink
the most populated clusters, dispatching some of its points in
neighboring cells. Figure~\ref{fig:geom_borders} exhibits the
influence of the $(d+1)^{\mathrm{th}}$ coordinate of the centroids 
on the position of the borders.

\subsection{Partial Balancing}
\label{sec:partial}
The proposed balancing strategy empirically converges towards 
clusters having the same size. Several stopping criteria can 
be applied, the most simple being a fixed maximum number of
iterations. It is also possible to target a particular value for
$\gamma$ which is recomputed at every step, either fixed or possibly
in proportion of the original imbalance factor. Early stopping the
balancing reduces the overall distortion of the Voronoi cells created
by the original $k$-means.

%% file: experiments.tex
\section{Experiments}
\label{sec:experiments}

\subsection{Datasets and Imbalance Factors Analysis}

Our analysis has been performed on descriptors extracted from a large
set of real-world images. We downloaded from Flickr one million images
to build the database and another set of one thousand images for the
queries.  Several description schemes were applied to these images,
namely SIFT local descriptors~\cite{Low04},
Bag-of-features~\cite{SiZ03} (BOF), GIST~\cite{OT01} and VLAD
descriptors~\cite{JDSP10}.  SIFT were
extracted from Hessian-Affine regions~\cite{MiS04} using the software
of~\cite{mikourl}.  The BOF vectors have been generated from these
local descriptors, using a codebook obtained by regular $k$-means
clustering with 1000 visual words.  The VLAD descriptors were
generated using a codebook of 64 visual words applied to the same SIFT
descriptors, leading to vectors of dimension~$64\times 128=8192$.  For
GIST, we have used the most common setup, i.e., the three color
channels and 3 scales, leading to 960-dimensional descriptors.
\smallskip

The global descriptors (BOF, GIST and VLAD) produce exactly one
descriptor per image, leading to one million vectors for each type of
descriptor.  In order to keep the same number of vectors for the SIFT
set, we have randomly subsampled the local descriptors to produce a
million-sized set. In all cases, we assume a closed-world setup, 
i.e., the dataset to be indexed is fixed, which is valid 
for most applications. 
\smallskip

\begin{table}[t]
\centering
\begin{tabular}{cccc}
\hline 
descriptor & dimensionality  &  \multicolumn{2}{c}{$\imf$}   \\
           &                 &  $k$=256   & $k$=1024 \\
\hline
SIFT     & 128  & 1.08 & 1.09  \\
BOF      & 1000 & 1.65 & 1.93 \\
GIST     & 960  & 1.72 & 3.75 \\
VLAD     & 8192 & 5.41 & 6.23 \\
\hline
\end{tabular}
\caption{Imbalance factor for $k$-means clustered state-of-the-art 
descriptors, measured on a dataset of one million images 
for two values of $k$. 
}
\label{tab:if}
\end{table}

Table~\ref{tab:if} reports the imbalance factors obtained for each
type of descriptors after performing a standard $k$-means clustering
on our database. It can be observed that higher dimensional vectors
tend to produce higher imbalance factors. BOF descriptors have an
imbalance factor which is lower than GIST for a comparable dimension,
which might be due to their higher sparsity.  Note that the value of~$k$
has a significant impact on~$\imf$: larger values of~$k$ lead to
significantly higher $\imf$ ($k$=256 and $k$=1024).  The low
values for $k$ we have considered here probably explain why $\imf$
measured for the SIFT descriptors in Table~\ref{tab:if} are lower than
those of the literature: Jegou \emph{et al.}~\cite{JDS10a} report $1.21$
and $1.34$ for codebooks of size $k$=20\,000 and $k$=200\,000,
respectively.
\smallskip

Our balancing strategy is especially interesting for global descriptors
for which, in contrast to local descriptors, exactly one query vector
is used. In this case, with perfectly balanced clusters, 
querying an image is performed in constant time. 
This is the rationale for focusing
our analysis on the well known BOF vectors.

\subsection{Evaluation of the proposed method} 

In this subsection we analyze the impact of our method on selectivity, recall and 
variability of the response time. We also analyze the convergence properties of our method. 
The parameter~$\alpha$ is set to $\alpha$=0.01 in all our experiments. 
\smallskip

\begin{figure}[t]
\centering
\includegraphics[width=\linewidth]{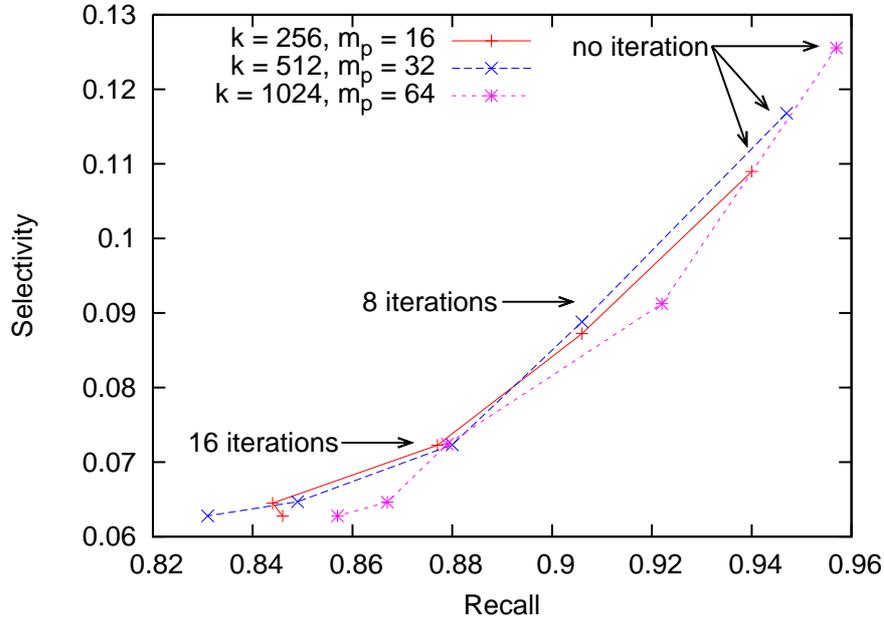}
\caption{Selectivity/recall performance: impact of partial and full balancing on this trade-off. 
For each value of $k$, the top-right points correspond to the original k-means partition (no iteration). 
From top to bottom, the points of a given curve correspond to $8$, $16$, $32$ and $64$ iterations. 
Similar to choosing a high value of $k$, our method reduces the selectivity (i.e., provides better efficiency) 
at the cost of lower recall. The different trade-off selectivity/recall are obtained, for our 
method, with a significantly lower variability of the response time. }
\label{fig:partial}
\end{figure}

\begin{figure}[t]
\centering
\includegraphics[width=\linewidth]{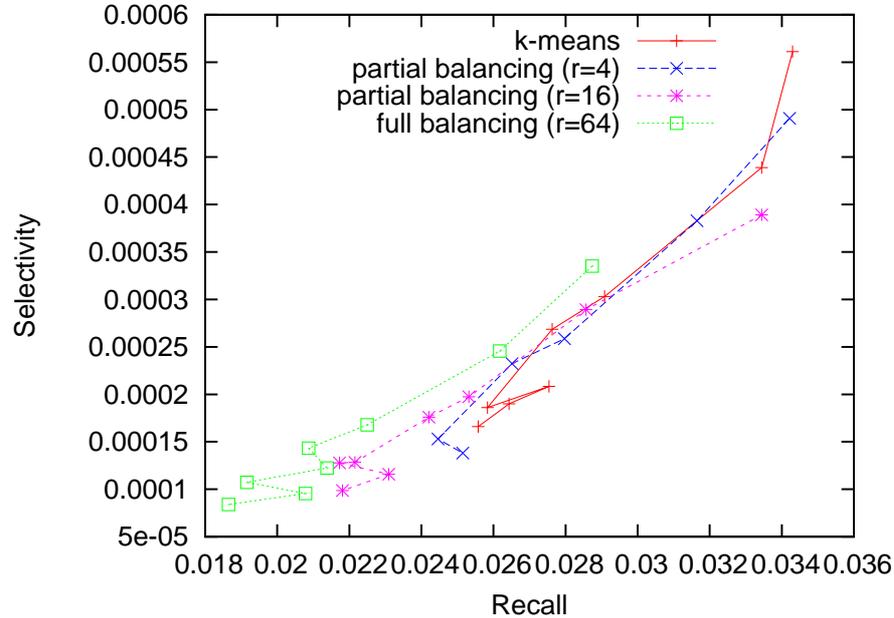}
\caption{Selectivity/recall performance: impact of partial and full balancing on this trade-off for $\ma=1$.
For each balancing strategy, the top-right points correspond to small values of $k$.
From top to bottom, $k$ varies between $256$ and $1024$.
Observe that if, for small values of $k$, balancing improves the performance in terms of this trade-off,
for large $k$, balancing tends to deteriorate this trade-off.}
\label{fig:partial_k_ma1}
\end{figure}

\begin{figure}[t]
\centering
\includegraphics[width=\linewidth]{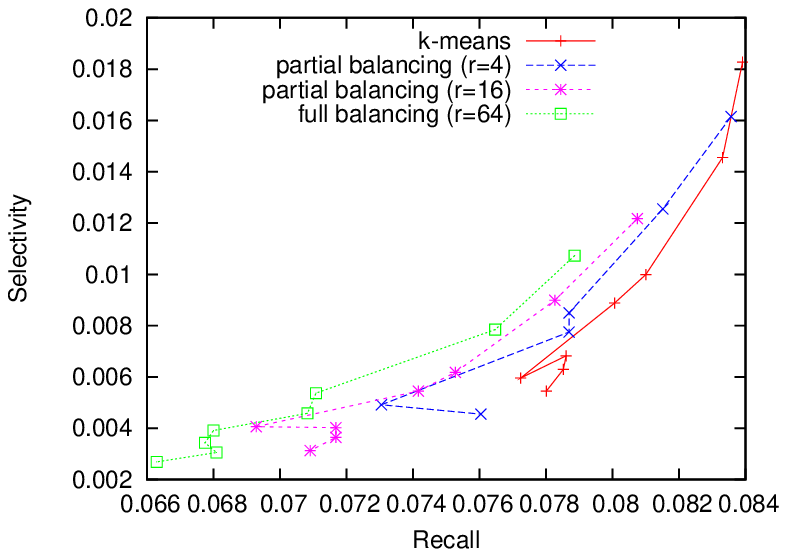}
\caption{Selectivity/recall performance: impact of partial and full balancing on this trade-off for $\ma=32$.
For each balancing strategy, the top-right points correspond to small values of $k$.
From top to bottom, $k$ varies between $256$ and $1024$.
In this case, balancing tend to deteriorate this trade-off.}
\label{fig:partial_k_ma32}
\end{figure}

{\noindent \bf Selectivity/recall performance:} 
Figure~\ref{fig:partial} shows the performance in terms of this trade-off 
for different values of~$k$. 
First note that the trade-off between selectivity and recall can be adjusted using 
the number~$k$ of clusters and the number~$\ma$ of probes. 
We keep the ratio $\ma/k$ constant in order 
to better show the impact of our method, which exhibits comparable performance 
with that of the k-means clustering in terms of selectivity and recall. 
Figures~\ref{fig:partial_k_ma1} and~\ref{fig:partial_k_ma32} shows comparable results when
$\ma$ is constant.
Note however that with our method a given selectivity/recall 
point is obtained with a much better (lower) 
variability of the response time, as shown later in this section. 
\smallskip

{\bf \noindent Impact of the number of iterations:}  
The number~$r$ of iterations performed by 
Equation~\ref{equ:update} is an important parameter of our method, 
as it controls to which extent complete balancing is enforced or not.
Figure~\ref{fig:partial} shows that
selectivity is reduced in the first iterations with a reasonable decrease of the recall, 
i.e., comparable to what we would obtain by modifying the number of clusters. 
The next iterations are comparatively less interesting, as the gain in
selectivity is obtained at the cost of a relatively higher decrease in
recall.  Modifying the stopping criterion allows
our method to attain a target imbalance factor which is
competitive with respect to the selectivity/recall trade-off. 
\smallskip

\begin{figure}[t]
\centering
\includegraphics[width=\linewidth]{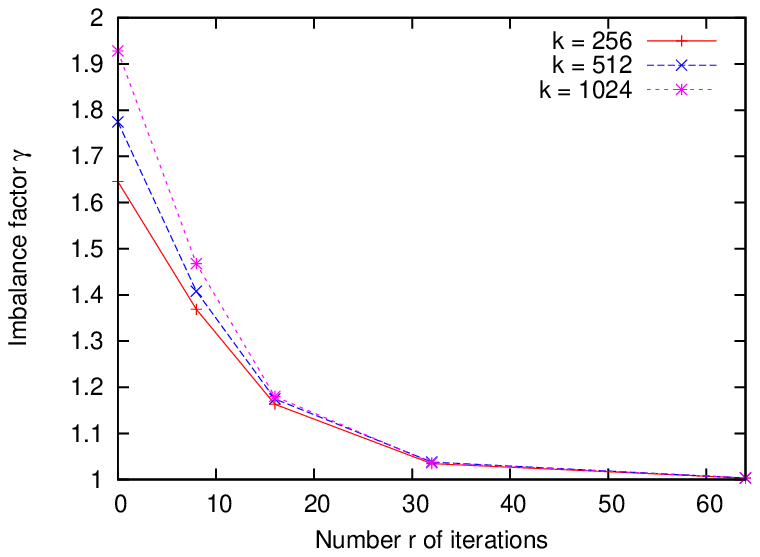}
\caption{Convergence speed}
\label{fig:convergence speed}
\end{figure}

{\noindent \bf Convergence speed:} 
Figure~\ref{fig:convergence speed} illustrates how the imbalance factor evolves along iterations. 
Only a few iterations are needed to attain reasonably balanced clusters. Our 
update procedure has a computational cost which is negligible compared with 
that of the clustering. 
Higher values of~$k$ do not require more iterations, which is somewhat surprising 
as more penalization terms have to be learned. 
Note that the convergence of our algorithm is not guaranteed, though in all the experiments
presented in this paper it has been observed.
\smallskip

\begin{figure}[t]
\centering
\includegraphics[width=\linewidth]{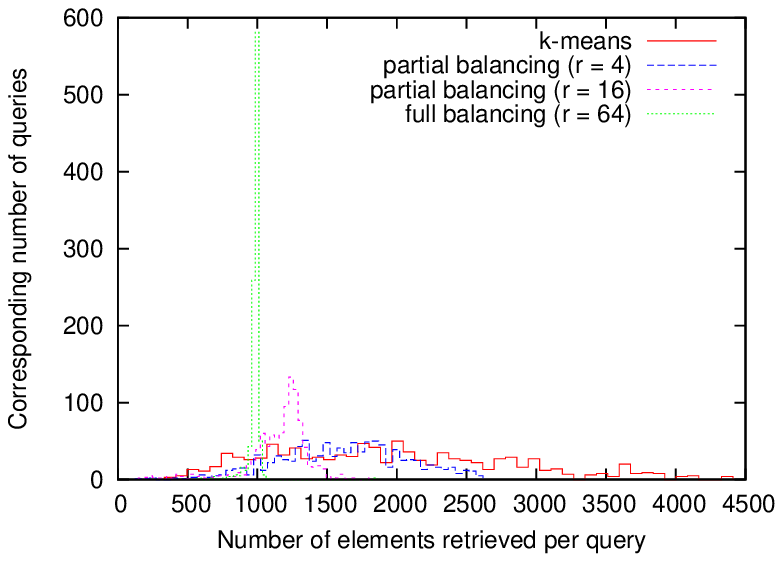}
\caption{Histograms of the number of elements returned, computed over our 1000 queries, 
for the original k-means and our algorithm with three number of iterations. 
Observe the tightness of the distribution in the case of our method, which reflects 
a very low variability in response time. }
\label{fig:histotime}
\vspace{-10pt}
\end{figure}

{\bf \noindent Variance of the query response time:}
The impact of our balancing strategy on the variability of the response time 
is illustrated by Figure~\ref{fig:histotime}, which gives the distribution 
of the number of elements returned by the indexing structure. 
The tight distribution obtained by our method shows 
that the objective of reducing the variability of the query time
resulting from unbalanced clusters is fulfilled: 
the response time is almost constant with full balancing. 
The partial balancing also leads to significantly improve the shape 
of the distribution, which has a significantly reduced 
variance compared with the original one.

\begin{figure}[t]
\centering
\includegraphics[width=\linewidth]{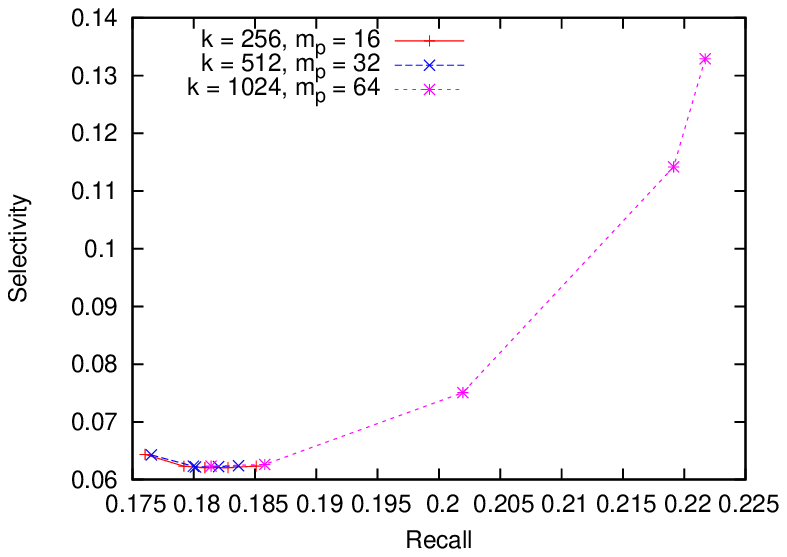}
\caption{Selectivity/recall performance: impact of partial and full 
balancing on this trade-off for Fisher kernel descriptors. From top to 
bottom, the points of a given curve correspond to $8$, $16$, $32$ and 
$64$ iterations.}
\label{fig:selrec-fisher}
\end{figure}

{\bf \noindent Impact of the choice of descriptors on observed results:}
In order to validate our approach on a different kind of descriptors, 
we tested it using Fisher kernels with $16$ gaussians. The query set 
is the concatenation of the Holidays dataset~\cite{JDS08} and the UKB one~\cite{NiS06}.
The results, as shown in figure~\ref{fig:selrec-fisher} are strongly 
dependent on the value of $k$. This is due to the fact that $k$-means 
clustering for small values of $k$ leads to well-balanced 
clusters ($\gamma \leq 1.1$) while $k=1024$ reaches an imbalance 
factor of $2.2$. In the latter case, balancing shows its efficiency in 
terms of selectivity, as expected.

\subsection{Is closed-world setup mandatory ?}
\begin{figure}[t]
\centering
\includegraphics[width=\linewidth]{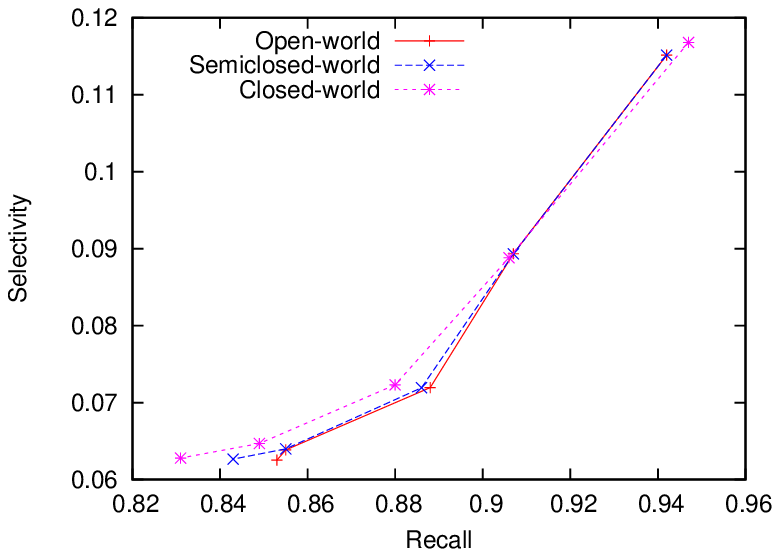}
\caption{Compared selectivity/performance for 3 different setups. 
Here, we used $k=512$ and $\ma=32$.}
\label{fig:closed-vs-open}
\end{figure}

Previous section presented results obtained in a closed-world
setup as it allows to achieve quasi-constant query time in all cases. 
However, figure~\ref{fig:closed-vs-open} shows that, as 
soon as distribution of the learning set is reasonably close to the 
one of the database, comparable selectivity-versus-recall compromise 
can be achieved in the open-world case. In this example, the database is the same as the 
one used in the previous experiments. For both closed-world and 
semiclosed-world setups, another 1 million images from Flickr are used 
as a learning set to train $k$-means. The different between both 
setups is that in the semiclosed-world one, balancing is learnt on the 
database itself while in the open-world setup, it is optimized on 
the learning set, which could lead to unbalanced database clusters.

Note nevertheless that quality of the balancing in semiclosed and 
open-world setups strongly depends on the learning set having comparable 
distribution to the one of the database. Therefore, their usage should 
be restricted to cases where this assumption is likely to be verified, 
as for example in cases where the learning set is a subset of the 
entire database.

%% file: conclusion.tex
\section{Conclusion}
\label{sec:conclusion}
Many high-dimensional indexing schemes rely on a partitioning of the
feature space into clusters obtained from a $k$-means type-of
approach. These schemes are efficient because they process a very
small number of cluster for answering each query. Their performance
suffer, however, from having to process clusters with very
different cardinalities since this causes great variations in the
response time to queries. This paper presents an algorithm that iteratively
balances clusters such that they become more equal in size. Reducing
the variance and the expectation of response times is a key issue when
targeting high-performance settings, especially when data has to be read from 
disk. Our 
experiments demonstrated that clusters are better balanced 
without significantly impacting the search quality. 
We are planning to index much
data collections where the imbalance factor will be higher, 
as for the promising VLAD descriptors~\cite{JDSP10}, 
increasing the need for a more uniform cluster distribution.